\pdfoutput=1

\documentclass[10pt,twocolumn,letterpaper]{article}

\usepackage[pagenumbers]{cvpr} 

\definecolor{cvprblue}{rgb}{0.21,0.49,0.74}
\usepackage[pagebackref,breaklinks,colorlinks,allcolors=cvprblue]{hyperref}
\usepackage{multirow}

\def\paperID{16369} 
\def\confName{CVPR}
\def\confYear{2025}

\title{EDM: Equirectangular Projection-Oriented Dense Kernelized Feature Matching}

\author{Dongki Jung\:\textsuperscript{\rm 1,2} \:\: Jaehoon Choi\:\textsuperscript{\rm 2} \:\: Yonghan Lee\:\textsuperscript{\rm 2} \:\: Somi Jeong\:\textsuperscript{\rm 1} \:\: Taejae Lee\:\textsuperscript{\rm 1} \\ Dinesh Manocha\:\textsuperscript{\rm 2} \:\:Suyong Yeon\:\textsuperscript{\rm 1}\\
$^{1}$NAVER LABS \:\: $^{2}$University of Maryland\\
{\tt\small jdk9405@umd.edu}
\vspace*{-4mm}
}

\begin{document}


\twocolumn[
{
    \renewcommand\twocolumn[1][]{#1}%
    \maketitle
    \centering
    \vspace{-4mm}
    \includegraphics[width=0.95\linewidth]{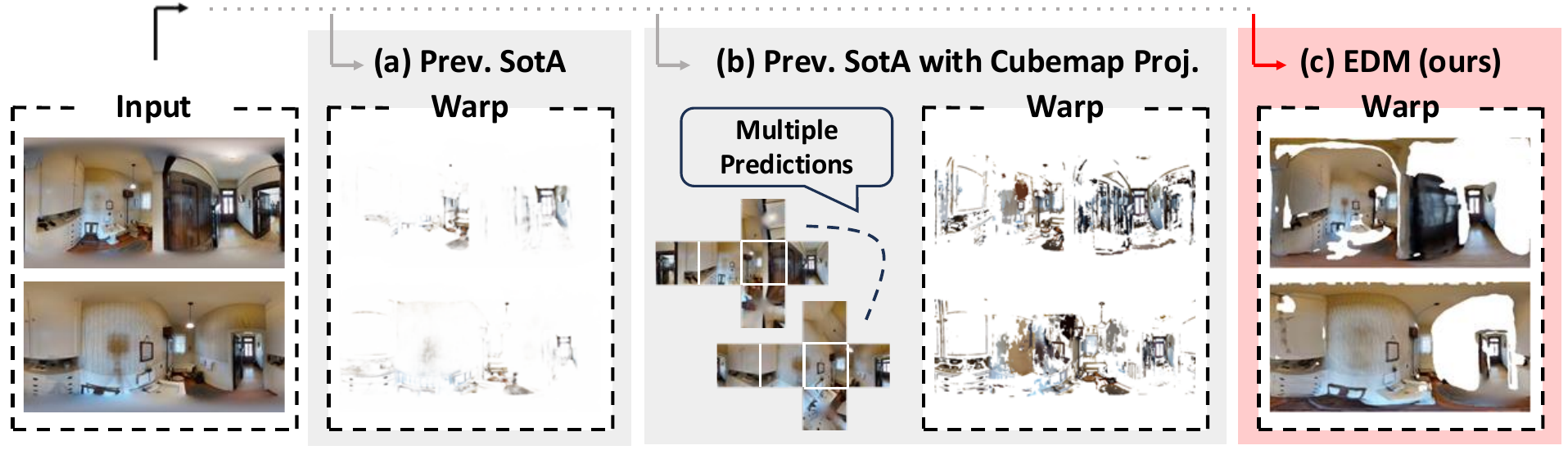}
    \vspace{-3mm}
    \captionof{figure}{ 
    (a) Previous state-of-the-art \citep{edstedt2023dkm} struggles to achieve accurate dense matching in equirectangular projection (ERP) images due to inherent distortions.
    (b) The ERP image can be transformed into a cubemap image, which consists of six perspective images. 
    However, this approach demands multiple independent iterations of inference for each pair of perspective images, increasing computational complexity and losing the global information in the ERP image.
    (c) Our proposed method, EDM, leverages the spherical camera model, rendering it robust against distortions.
    \textbf{Warp} refers to results obtained by multiplying the warped image with the predicted certainty map,
    demonstrating that our method yields more accurate dense matches.
    }
    \vspace*{-1mm}
    \label{fig:introduction}
}
]

\vspace*{5mm}
\begin{abstract}
  We introduce the first learning-based dense matching algorithm, termed Equirectangular Projection-Oriented Dense Kernelized Feature Matching (EDM), specifically designed for omnidirectional images. Equirectangular projection (ERP) images, with their large fields of view, are particularly suited for dense matching techniques that aim to establish comprehensive correspondences across images. However, ERP images are subject to significant distortions, which we address by leveraging the spherical camera model and geodesic flow refinement in the dense matching method. To further mitigate these distortions, we propose spherical positional embeddings based on 3D Cartesian coordinates of the feature grid. Additionally, our method incorporates bidirectional transformations between spherical and Cartesian coordinate systems during refinement, utilizing a unit sphere to improve matching performance. We demonstrate that our proposed method achieves notable performance enhancements, with improvements of +26.72 and +42.62 in AUC@5° on the Matterport3D and Stanford2D3D datasets. 
  Project Page: \url{https://jdk9405.github.io/EDM}
\end{abstract}
\vspace*{-2mm}
\section{Introduction} \label{section:introduction}
Omnidirectional images, also known as 360\textdegree\ images, provide significant advantages owing to their expansive fields of view, offering more contextual information and versatility ~\citep{xu2020state,zhang2023survey,matzen2017low,da20223d,guerrero2020s}. 
These spherical images enable a comprehensive representation of environments, facilitating a deeper understanding of spatial information. Their utility extends to aiding robot navigation \citep{winters2000omni,menegatti2004image} and autonomous vehicle driving \citep{pandey2011ford} by minimizing blind spots.
360\textdegree\ images also can be utilized in a diverse range of applications, from creating immersive AR/VR experiences to practical uses in interior design \citep{amalia2023case}, tourism \citep{saurer2010omnitour}, and real estate photography \citep{chang2017matterport3d}. Integrating omnidirectional images into virtual house tours allows customers to experience an immersive view, enabling them to fully engage themselves in the service.
Moreover, the adoption of omnidirectional images contributes to more efficient data collection. 
By replacing the need for multiple perspective images, omnidirectional images can reduce both the cost and time associated with data scanning.
%
The large field of view provided by 360\textdegree\ images has also demonstrated superiority over narrower views in 3D motion estimation \citep{nelson1988finding, lee2000large, fermuller2001geometry}.

Feature matching plays a critical role in numerous 3D computer vision tasks, including mapping and localization. 
Traditionally, Structure from Motion (SfM) \citep{schonberger2016structure} leverages feature matching to estimate relative poses.
Recent advancements have introduced semi-dense or dense approaches for feature matching such as LoFTR \citep{sun2021loftr} and DKM \citep{edstedt2023dkm}, which demonstrate superior performance in repetitive or textureless environments compared to keypoint-based methods \citep{lowe2004distinctive,rublee2011orb,detone2018superpoint,sarlin2020superglue,li2022decoupling}. 
These methods have been mainly developed for perspective 2D images and videos, but encounter challenges when applied to omnidirectional images.
For example, to adapt matching methods for spherical images, two prevalent approaches for sphere-to-plane projections are the equirectangular projection (ERP) and the cubemap projection \citep{xu2020state}.
ERP images exhibit significant distortions, particularly near the pole regions, which hinder the effective application of perspective methods.
On the other hand, the cubemap format, consisting of six perspective images, can be processed independently without such distortions.
However, this approach involves the costly computation of multiple inferences for each pair of perspective images, resulting in the loss of global information from a single spherical image and diminishing feature matching capabilities due to the reduced field of view in each perspective image.
These challenges are shown in Fig. \ref{fig:introduction} (a) and (b).

\paragraph{Main Results} 
In this paper, we propose EDM, a distortion-aware dense feature matching method for omnidirectional images, addressing challenges that existing detector-free approaches \citep{sun2021loftr,edstedt2023dkm,edstedt2023roma} struggle to overcome. 
To the best of our knowledge, EDM is the first learning-based method designed for dense matching and relative pose estimation between two omnidirectional images. 
As seen in Fig. \ref{fig:introduction}, our method defines feature matching in 3D coordinates, specifically addressing the challenges posed by distortions of ERP images. 
We accomplish this based on the integration of two novel steps: a Spherical Spatial Alignment Module (SSAM) and specific enhancements in Geodesic Flow Refinement.
The SSAM leverages spherical positional embeddings for ERP images and incorporates a decoder to generate the global matches.
Furthermore, the Geodesic Flow Refinement step employs coordinate transformation to refine the residuals of correspondences.
Compared to both recent sparse and dense feature matching methods \citep{zhao2015sphorb,gava2023sphereglue,edstedt2023dkm,edstedt2023roma}, our approach results in significant performance improvement of +26.72 and +42.62 AUC@5\textdegree\ in relative pose estimation for spherical images on the Matterport3D \citep{chang2017matterport3d} and Stanford2D3D \citep{armeni2017joint} datasets.
Additionally, we evaluate our method qualitatively on the EgoNeRF \citep{choi2023balanced} and OmniPhotos \citep{bertel2020omniphotos} datasets, demonstrating robust performance across diverse environments.
The main contributions of this paper are summarized as follows:
\begin{itemize}
    \item We introduce a novel approach for estimating dense matching across ERP images using geodesic flow on a unit sphere.
    \item We propose a Spherical Spatial Alignment Module that utilizes Gaussian Process regression and spherical positional embeddings to establish 3D correspondences between omnidirectional images. In addition, we use Geodesic Flow Refinement by enabling conversions between coordinates to refine the displacement on the surface of the sphere.
    \item With azimuth rotation for data augmentation, we achieve state-of-the-art performance in dense matching and relative pose estimation between two omnidirectional images.
\end{itemize}

\section{Related Work}
\paragraph{Omnidirectional Images}
The popularity of consumer-level 360\textdegree\ cameras has led to increased interest in spherical images, which offer comprehensive coverage of the field of view from a single vantage point. 
These images are often represented using equirectangular projection (ERP) \citep{xu2020state}, facilitating their utilization in various computer vision tasks. Recent advancements in computer vision have leveraged ERP images for diverse tasks such as object detection \citep{coors2018spherenet, su2017learning}, semantic segmentation \citep{jiang2019spherical, zhang2019orientation}, depth estimation \citep{jiang2021unifuse, wang2020bifuse, shen2022panoformer, li2022omnifusion, rey2022360monodepth, li2021panodepth, yun2022improving}, omnidirectional Simultaneous Localization and Mapping \citep{won2020omnislam}, scene understanding \citep{sun2021hohonet}, and neural rendering \citep{choi2024omnilocalrf, kim2024omnisdf, ma2024fastscene, li2024omnigs}.

Despite the utility of ERP images, their unique geometry presents several challenges in visual representation.
As ERP images are obtained through projecting a sphere onto a plane, a single spherical image can be expressed by multiple distinct ERP images.
Additionally, ensuring perfect alignment of their left and right extremities is essential. 
While some research methods have introduced rotation-equivariant convolutions \citep{cohen2018spherical, esteves2018learning}  to address these issues, their implementation often demands increased computational resources. 
To mitigate this constraint, we propose an azimuth rotation approach for data augmentation, under the assumption that maintaining the downward orientation of scanned omnidirectional images parallel to gravity offers benefits \citep{bergmann2021gravity}.

\paragraph{Feature Matching}
Local feature matching has relied on detector-based methods, encompassing both traditional hand-crafted techniques \citep{lowe2004distinctive, rublee2011orb} and learning-based approaches \citep{detone2018superpoint, revaud2019r2d2, li2022decoupling, liu2019gift, tyszkiewicz2020disk}.
These methods typically involve detecting keypoints, computing descriptor distances between paired keypoints, and performing matching via mutual nearest neighbor search. 
SuperGlue \citep{sarlin2020superglue} introduces a learning-based paradigm, optimizing visual descriptors using an attentional graph neural network and an optimal matching layer. 
However, detector-based methods face limitations in terms of accurately detecting keypoints, particularly in repetitive or indiscriminative regions. 
In contrast, detector-free or dense methods \citep{sun2021loftr, melekhov2019dgc, truong2020glu, truong2021learning, edstedt2023dkm, edstedt2023roma} offer a solution to the keypoint detection issue, providing dense feature matches at the pixel level.

While the aforementioned methods are tailored for perspective images, they often fail to address the unique challenges of spherical cameras. 
SPHORB \citep{zhao2015sphorb}, an extension of ORB \citep{rublee2011orb}, mitigates distortion in ERP images using a geodesic grid and local planar approximation \citep{eder2020tangent}. 
Similarly, learning-based matching methods such as SphereGlue \citep{gava2023sphereglue, gava2024spherecraft} and PanoPoint \citep{zhang2023panopoint} adapt keypoint matching techniques for spherical imagery. 
CoVisPose \citep{hutchcroft2022covispose, nejatishahidin2023graph} explores layout features for estimating camera poses over large baselines yet remains constrained by detected feature information. 
Therefore, we propose a novel dense matching method that extracts all matches without keypoint detection in spherical images.



\section{Preliminaries} \label{section:preliminaries}
\subsection{Spherical and Cartesian Coordinate}
\begin{figure}[h]
    \centering
    \vspace*{-3mm}
    \includegraphics[width=0.85\linewidth]{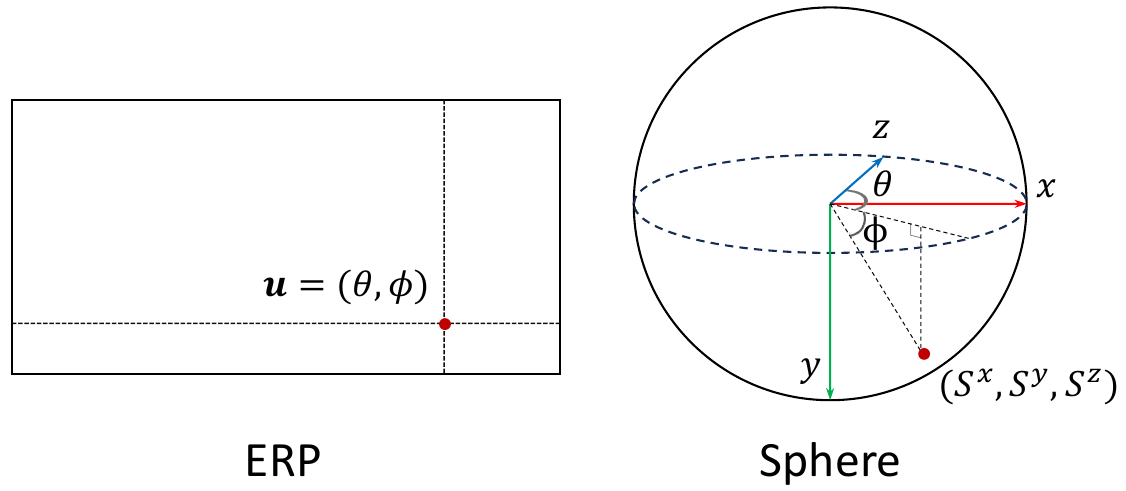}
    \caption{Coordinate system.}
    \vspace*{-4mm}
    \label{fig:coord}
    \vspace*{-5mm}
\end{figure}


\begin{equation}
        \label{projection}
        \left\{
        \begin{aligned}
            &S^x = \sin(\theta)\cos(\phi) \\
            &S^y = \sin(\phi) \\
            &S^z = \cos(\theta)\cos(\phi)
        \end{aligned}
        \right.
        \quad
        \left\{
        \begin{aligned}
            &\theta = \text{arctan}(\frac{S^x}{S^z}) \\ 
            &\phi = \text{arcsin}(\frac{S^y}{|\mathbf{S}|})
        \end{aligned}
        \right.
\end{equation}
Although ERP images are displayed in 2D space, they actually represent a collection of flattened rays normalized to a unit scale within a spherical camera model.
Thus, we can express the coordinate conversion equation $\mathbf{u} = \boldsymbol{\pi}(\mathbf{S})$ between the spherical coordinates $\mathbf{u}=(\theta, \phi)$ and the 3D Cartesian coordinates $\mathbf{S} = (S^x, S^y, S^z)$ as shown in Fig. \ref{fig:coord}.
Each value of $\theta \in [-\pi, \pi]$ and $\phi \in [-\frac{\pi}{2}, \frac{\pi}{2}]$ indicates the longitude and latitude. We utilize this coordinate transformation $\boldsymbol{\pi}(\cdot)$ in Section \ref{section:ssam} and Section \ref{section:gfr} to handle the spherical camera model effectively.



\subsection{Dense Kernelized Feature Matching}
\label{section:dense_matching}
Dense matching is the task of finding dense correspondence and estimating 3D geometry from two images $\left( I_\mathcal{A}, I_\mathcal{B} \right)$. 
%
%
Recently, DKM \citep{edstedt2023dkm} introduced a kernelized global matcher and warp refinement, formulating this problem as finding a mapping $f \to \mathbf{u}$ where $\mathbf{u}$ are 2D spatial coordinates.
First, DKM extracts multi-scale features using a ResNet50 encoder \citep{he2016deep},
\begin{equation} \label{encoding}
    \{f_{\mathcal{A}}^{l}\}_{l=1}^{L} = \text{Encoder}(I_{\mathcal{A}}), \quad \{f_{\mathcal{B}}^{l}\}_{l=1}^{L} = \text{Encoder}(I_{\mathcal{B}}),
\end{equation}
where the strides are defined as elements of the set $l \in \{2^{0},...,2^{L-1}\}$. Coarse features are associated with stride $\{32,16\}$, and fine features correspond to $\{8,4,2,1\}$. 

At the coarse level, it consists of a kernelized regression to estimate the posterior mean $\mu_{\mathcal{A} | \mathcal{B}}$ using a Gaussian Process (GP) formulation. 
GP regression generates a probabilistic distribution using the feature information conditioned on frame $\mathcal{B}$ to estimate coarse global matches.
The normalized 2D feature grid $f_B^{\text{grid}} \in \mathbb{R}^{h \times w \times 2}$, where $h$ and $w$ denote the resolution of the feature grid, is embedded into $\chi_{\mathcal{B}}$ with an additional cosine embedding \citep{snippe1992discrimination} to induce multimodality in GP. The embedded coordinates are processed by an exponential cosine similarity kernel $K$ to calculate $\mu_{\mathcal{A} | \mathcal{B}}$,
\begin{equation} \label{mu}
\begin{aligned}
    \mu_{\mathcal{A} | \mathcal{B}} &= K_{\mathcal{AB}}(K_{\mathcal{BB}} + \sigma_n^2 I)^{-1}\chi^{\text{coarse}}_{\mathcal{B}}, \\   
\end{aligned}
\end{equation}
\begin{equation}
\left\{
\begin{aligned}
    &K_{mn} = \exp \left( \tau \left( \frac{ f_m \boldsymbol{\cdot} f_n }{\sqrt{(f_m \boldsymbol{\cdot} f_m)  (f_n \boldsymbol{\cdot} f_n) + \varepsilon}} - 1 \right) \right), \\
    &\chi^{\text{coarse}}_{\mathcal{B}} = \text{cos}(Wf_{\mathcal{B}}^\text{grid} + b), \\
\end{aligned}
\right.
\end{equation}
where $\tau = 5$, $\epsilon=10^{-6}$, and the standard deviation of the measurement noise $\sigma_n = 0.1$ in the experiments. $W$ and $b$ are the weights and biases of a $1\times1$ convolution layer. 
Then, CNN embedding decoder \citep{yu2018learning} yields the initial global matches $\hat{\mathbf{u}}^{\text{coarse}}_{\mathcal{A} \to \mathcal{B}}$ and confidence of matches $\hat{c}^{\ \text{coarse}}_{\mathcal{A} \to \mathcal{B}}$ from the concatenation of the reshaped estimated posterior mean $\mu_{\mathcal{A} | \mathcal{B}}^{\text{grid}}$ and the coarse features,
\begin{equation} \label{decoder}
\begin{aligned}
    \left( \hat{\mathbf{u}}_{\mathcal{A} \to \mathcal{B}}^{\text{coarse}}, \hat{c}_{\mathcal{A} \to \mathcal{B}}^{\ \text{coarse}} \right) &= \text{Decoder}(\mu_{\mathcal{A} | \mathcal{B}}^{\text{grid}} \oplus f_\mathcal{A}^{\text{coarse}}).
\end{aligned}
\end{equation}
At the fine level, the warp refiners estimate the residual displacement using the previous matches and feature information. 
The process is described as follows,
\begin{equation} \label{refine}
\begin{aligned}
    & \quad \left( \triangle\hat{\mathbf{u}}_{\mathcal{A} \to \mathcal{B}}^{l+1}, \ \triangle\hat{c}_{\mathcal{A} \to \mathcal{B}}^{\ l+1} \right) \\
    \small= \text{Refiner}^{l+1} & \scriptstyle \left( f_\mathcal{A}^{l+1} \oplus f_{\mathcal{B} \to \mathcal{A}}^{l+1} \oplus Corr_{\Omega_k}^{l+1} \oplus \hat{\mathbf{u}}_{\mathcal{A} \to \mathcal{B}}^{l+1} - \mathbf{u}_{\mathcal{A}}^{l+1} \right), \\
\end{aligned}
\end{equation}
\begin{equation}\label{refine_subeq}
\left\{
\begin{aligned}
    & f_{\mathcal{B} \to \mathcal{A}}^{l+1} = f_{\mathcal{B}}\langle \hat{\bf{u}}_{\mathcal{A} \to \mathcal{B}}^{l+1} \rangle, \\
    & f_{\mathcal{B} \to \mathcal{A}, \ \Omega_{k}}^{l+1} = f_{\mathcal{B}}\langle \Omega_{k},(\hat{\bf{u}}_{\mathcal{A} \to \mathcal{B}}^{l+1}) \rangle, \\
    & Corr_{\Omega_k}^{l+1} = \sum_{\text{channel}} f_{\mathcal{A}}^{l+1} \ f_{\mathcal{B} \to \mathcal{A}, \ \Omega_k}^{l+1},
\end{aligned}
\right.
\end{equation}
where $\Omega_{k}(\mathbf{u}) = \mathbf{u} + \mathbf{p} \ (\|\mathbf{p}\|_\infty \leq k)$ is the patch sized $k$, $\langle \cdot \rangle$ means the bilinear interpolation function, $Corr_{\Omega_k}^{l+1}$ represents local correlation between the features, and $\textbf{u}_{\mathcal{A}}^{l+1}$ indicates the grid in $f_{\mathcal{A}}^{l+1}$.
Finally, it recursively updates the matching points and confidence by adding the residuals to the previous information and upsampling until reaching the same resolution as the input images,
\begin{equation}\label{plus_upsample}
\begin{aligned}
    \hat{\mathbf{u}}^{l}_{\mathcal{A} \to \mathcal{B}} &= \hat{\mathbf{u}}^{l+1}_{\mathcal{A} \to \mathcal{B}} + \triangle\hat{\mathbf{u}}^{l+1}_{\mathcal{A} \to \mathcal{B}}, \\
    \hat{c}^{\ l}_{\mathcal{A} \to \mathcal{B}} &= \hat{c}^{\ l+1}_{\mathcal{A} \to \mathcal{B}} + \triangle\hat{c}^{\ l+1}_{\mathcal{A} \to \mathcal{B}}.
\end{aligned}
\end{equation}

\begin{figure}[t]
    \centering
    \includegraphics[width=1.0\linewidth]{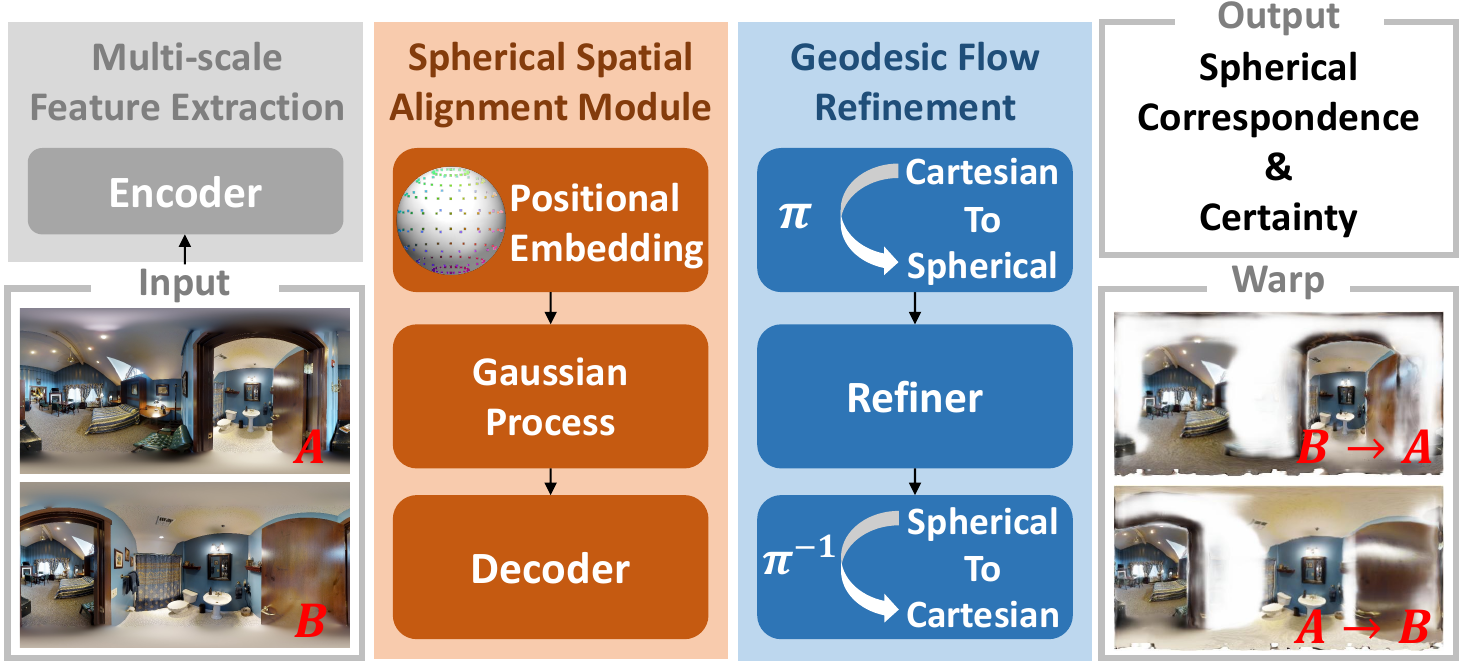}
    \vspace*{-6mm}
    \caption{
    \label{fig:architecture} Overview of our approach. It consists of three steps: Multi-scale Feature Extraction, Spherical Spatial Alignment Module (Sec. \ref{section:ssam}), and Geodesic Flow Refinement (Sec. \ref{section:gfr}).
    }
\end{figure}

\section{Our Proposed Method} \label{section:method}
The overall process is illustrated in Fig. \ref{fig:architecture}.
Following the approach outlined in Section \ref{section:dense_matching}, we first utilize ERP images $I_{\mathcal{A}}$ and $I_{\mathcal{B}}$ as input and extract multi-scale features $f_{\mathcal{A}}$ and $f_{\mathcal{B}}$.
Different from \citep{edstedt2023dkm}, we reformulate the problem as finding a mapping $f \to \mathbf{S}$ using 3D Cartesian coordinates.
%
%
We introduce the Spherical Spatial Alignment Module, a global matcher utilizing a spherical camera system to compensate for distortions caused by sphere-to-plane projection in ERP images.
We then formalize the geodesic flow on a unit sphere and establish projections between equirectangular and spherical spaces to refine matches. 
In addition, to enhance the robust accuracy of our method, we leverage randomized azimuth rotation during the training process.
\subsection{Spherical Spatial Alignment Module}\label{section:ssam}

\begin{figure}[t]
    \centering
    \includegraphics[width=1.0\linewidth]{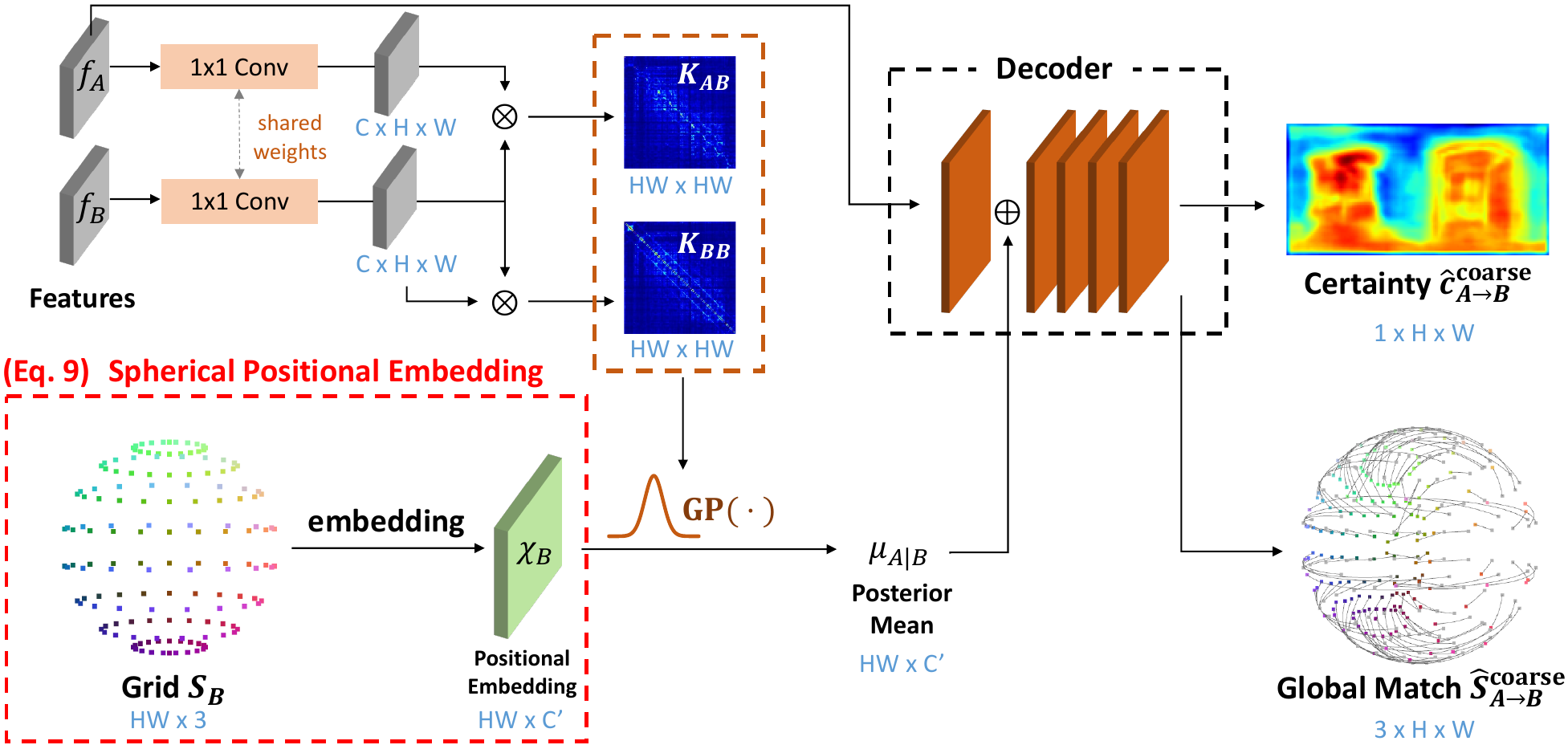}
    \caption{
    Our Spherical Spatial Alignment Module.
    We present Spherical Positional Embedding (red dotted box). 
    %
    %
    The embedding decoder generates the global matches $\hat{\mathbf{S}}^{\text{coarse}}_{{A} \to {B}}$.
    Here, the gray curved lines represent the geodesic flow between $\mathbf{S}_{A}$ and $\mathbf{S}_{B}$.
    %
    $\oplus$ denotes concatenation, $\otimes$ means reshape and matrix multiplication.  We provide the matrix dimensions of intermediate features for reference.
    }
    \label{fig:ssap}
\end{figure}

Our Spherical Spatial Alignment Module (SSAM) conducts global matching at a coarse level through Gaussian Process (GP) regression, depicted in Fig. \ref{fig:ssap}.
GP predicts the posterior mean $\mu_{\mathcal{A} | \mathcal{B}}$ from the embeddings as in Eq. \ref{mu}. 
%
Due to the pronounced distortions in the polar regions of ERP images, spherical positional embedding/encoding is frequently employed to mitigate this challenge \citep{chen2022text2light,li2023mathcal,li2023panoramic}.
%
Here, we explicitly apply positional embeddings with 3D Cartesian coordinates, derived from the 2D spherical feature grid and the inverse transformation function $\boldsymbol{\pi}^{-1}(\cdot)$,
\begin{equation} \label{positional_embedding}
    \chi_{\mathcal{B}}^{\text{coarse}} = \text{cos}(W\boldsymbol{\pi}^{-1}(f_{\mathcal{B}}^{\text{grid}})+b).
\end{equation}
Our proposed positional embedding facilitates the utilization of embedded coordinates $\chi_{\mathcal{B}}^{\text{coarse}}$ to promote distortion awareness within the ERP images.
Additionally, this embedding ensures structural consistency along the boundaries of ERP images by leveraging relative spatial information within the 3D Cartesian grid.
The outputs of the subsequent embedding decoder provide the initial global matches $\hat{\mathbf{S}}_{\mathcal{A} \to \mathcal{B}}^{\text{coarse}}$ on the unit sphere and the ERP certainty map $\hat{c}_{\mathcal{A} \to \mathcal{B}}^{\text{coarse}}$,
\begin{equation} \label{matcher}
    \left( \hat{\mathbf{S}}_{\mathcal{A} \to \mathcal{B}}^{\text{coarse}}, \hat{c}_{\mathcal{A} \to \mathcal{B}}^{\text{coarse}} \right) = \text{Decoder}(\mu_{\mathcal{A} | \mathcal{B}}^{\text{}} \oplus f_\mathcal{A}^{\text{coarse}}).
\end{equation}

\subsection{Geodesic Flow Refinement}\label{section:gfr}

\begin{figure}[t]
    \centering
    \includegraphics[width=1.0\linewidth]{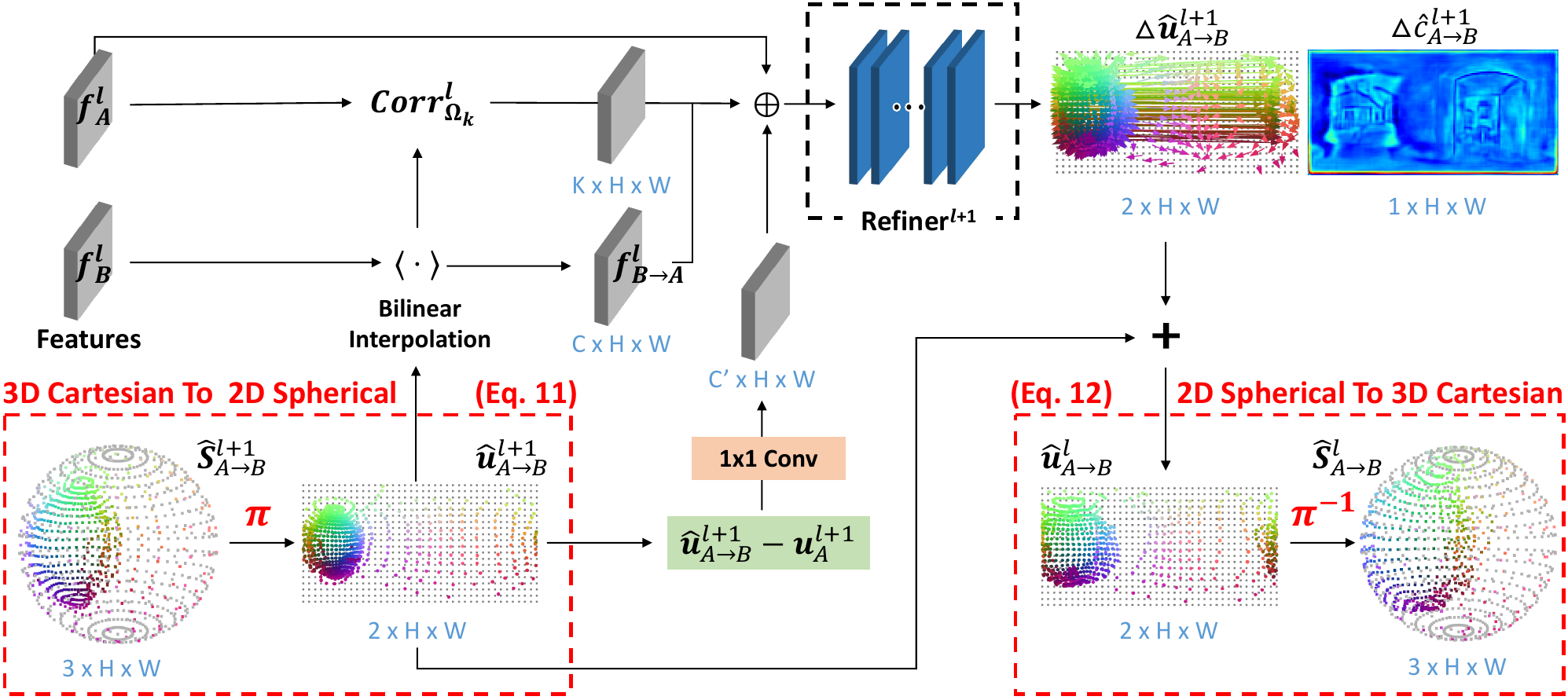}
    \caption{
    Our proposed Geodesic Flow Refinement.
    Refining the displacement along curved lines on the spherical surface presents significant challenges. To address this, we project the displacement into the ERP space for refinement (Cartesian to spherical) and subsequently unproject it back onto the spherical surface for further refinement (spherical to Cartesian).
    }
    \label{fig:refine} 
\end{figure}

%
%
In our SSAM approach, as the geodesic flow must reside on the unit sphere, directly defining warp refinement on the surface of the sphere makes it impossible to update the residuals linearly. 
Thus, we circumvent this problem by enabling a conversion between the 3D Cartesian coordinates and the 2D equirectangular space, as illustrated in Fig. \ref{fig:refine},
\begin{equation} \label{proj_}
\begin{aligned}
    %
    \mathbf{\hat{u}}^{l+1}_{\mathcal{A} \to \mathcal{B}} = \boldsymbol{\pi}(\mathbf{\hat{S}}^{l+1}_{\mathcal{A} \to \mathcal{B}}).
\end{aligned}
\end{equation}
After following all the processes outlined in Eq. \ref{refine} for refinement, we update the residuals as described in Eq. \ref{plus_upsample}.
As this refinement stage iterates repeatedly, the predicted $\hat{\mathbf{u}}_{\mathcal{A} \to \mathcal{B}}^{l}$ is back-projected into 3D Cartesian coordinates,
\begin{equation} \label{unproj_}
\begin{aligned}
    %
    \mathbf{\hat{S}}^{l}_{\mathcal{A} \to \mathcal{B}} = \boldsymbol{\pi}^{-1}(\mathbf{\hat{u}}^{l}_{\mathcal{A} \to \mathcal{B}}).
\end{aligned}
\end{equation}

\subsection{Augmentation}
\begin{figure}[t]
    \centering
    \includegraphics[width=1.0\linewidth]{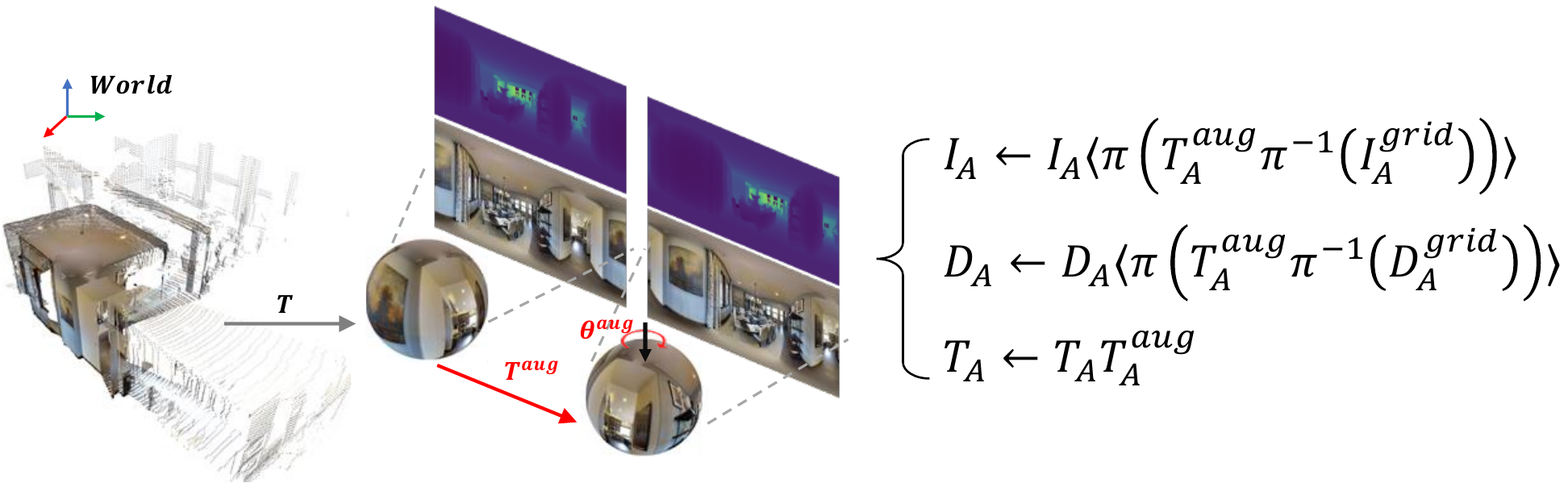}
    \caption{
        Maintaining consistent geometry, ERP can produce multiple visual representations based on $\theta^{\text{aug}}$.
        }
    \label{fig:augmentation}
\end{figure}

A single omnidirectional image can be transformed into multiple distinct ERP images, as shown in Fig. \ref{fig:augmentation}.
This transformation is feasible by capturing the full spectrum of rays and ensuring a seamless representation in the spherical input image, which facilitates the generation of diverse ERP images while maintaining consistent geometric properties in the world space.
Consequently, we define a horizontal rotation matrix $T_{\mathcal{A}}^{\text{aug}}$ with a randomly selected azimuth angle $\theta_{\mathcal{A}}^{\text{aug}} \in [0, 2\pi]$ during training. 
Based on $T_{\mathcal{A}}^\text{aug}$, we rotate and redefine the ERP image $I_{\mathcal{A}}$, the depth map $D_{\mathcal{A}}$, and the pose $T_{\mathcal{A}}$.
Notably, this transformation adjusts $T_{\mathcal{A}}$ and $D_{\mathcal{A}}$ together, ensuring consistent geometry in the world space.
The same process is applied to the counterpart frame $\mathcal{B}$.

\subsection{Loss}
Utilizing dense ground truth depth maps and aligned camera poses, we can derive ERP depth $D_{\mathcal{A} \to \mathcal{B}}$ and matches $\mathbf{S}_{\mathcal{A} \to \mathcal{B}}$ during the warping process from frame $\mathcal{A}$ to $\mathcal{B}$ within the spherical coordinate system.
We adopt the certainty estimation method proposed by \citet{edstedt2023dkm}, which involves finding consistent matches using relative depth consistency between frames $\mathcal{A}$ and $\mathcal{B}$,
\begin{equation} \label{certainty}
    c_{\mathcal{A} \to \mathcal{B}} = \left|\frac{D_{\mathcal{A} \to \mathcal{B}}-D_{\mathcal{B}}}{D_{\mathcal{B}}}\right| < \alpha,
\end{equation}
where $\alpha$ is 0.05. The binary mask $c_{\mathcal{A} \to \mathcal{B}}$ represents the ground truth certainty map.
%
%
Diverging from the approach outlined in \citet{edstedt2023dkm}, our method constrains the predicted matches $\mathbf{\hat{S}}^{l}_{\mathcal{A} \to \mathcal{B}}$, composed of 3D Cartesian coordinates, to reside on the surface of the unit sphere.
This implies that the predicted matches can be interpreted as the ray directions of the spherical camera.
Instead of defining the loss function based on the Euclidean distance between the predicted matches $\mathbf{\hat{S}}^{l}_{\mathcal{A} \to \mathcal{B}}$ and the ground truth matches $\mathbf{S}^{l}_{\mathcal{A} \to \mathcal{B}}$, we use the angular difference between the ray directions. 
Consequently, this approach ensures that $\mathbf{\hat{S}}^{l}_{\mathcal{A} \to \mathcal{B}}$ is optimized along the surface of the unit sphere.
We define our regression loss $L_{\text{r}}^{l}$ using cosine similarity to measure the angular difference.
For the certainty loss $L_{\text{c}}^{l}$, we employ the binary cross-entropy function, as utilized in \citet{edstedt2023dkm},
\begin{equation} \label{main_loss}
    \mathcal{L}_{\text{r}}^{l} = \sum_{\text{grid}} c^{l}_{\mathcal{A} \to \mathcal{B}} \odot (1 - \frac{\lVert \mathbf{S}_{\mathcal{A} \to \mathcal{B}}^{l} \boldsymbol{\cdot} \mathbf{\hat{S}}_{\mathcal{A} \to \mathcal{B}}^{l}  \rVert}{\lVert \mathbf{S}_{\mathcal{A} \to \mathcal{B}}^{l} \rVert \lVert \mathbf{\hat{S}}_{\mathcal{A} \to \mathcal{B}}^{l} \rVert}),
\end{equation}

\begin{equation}
    \label{conf_loss}
    \mathcal{L}_{\text{c}}^{l} = \sum_{\text{grid}} c^{l}_{\mathcal{A} \to \mathcal{B}} \text{log} \hat{c}^{l}_{\mathcal{A} \to \mathcal{B}} + (1 - c^{l}_{\mathcal{A} \to \mathcal{B}}) \text{log} (1 - \hat{c}^{l}_{\mathcal{A} \to \mathcal{B}}).
\end{equation} 
%
The total loss function comprises a weighted sum of the regression loss and the certainty loss, as detailed in \citet{zhou2021patch2pix,melekhov2019dgc,tan2022eco,edstedt2023dkm}, with $\lambda$ set at 0.01,
\vspace*{-2mm}
\begin{equation}\label{eq:total_loss}
\begin{aligned}
    L_{\text{total}} = \sum_{l=1}^{L} L_{\text{r}}^{l} + \lambda L_{\text{c}}^{l}. \\
\end{aligned}
\end{equation}

\section{Experiments} \label{section:experiments}
\vspace*{-2mm}
\subsection{Experiments Settings}

\paragraph{Matterport3D Dataset}
Training our method requires ERP input images, ground truth depth maps, and aligned poses.
The Matterport3D dataset \citep{chang2017matterport3d} encompasses 90 indoor scenes represented by 10,800 panoramas reconstructed as textured meshes.
However, the dataset lacks pose and depth information for \textit{skybox} images, which are essential for creating ERP images. 
Previous works have addressed this limitation by rendering both images and depth maps from the textured mesh \citep{zioulis2018omnidepth} or by employing 360\textdegree\ SfM to estimate poses \citep{rey2022360monodepth}.
In our approach, we generate the poses for \textit{skybox} images directly from the originally proposed camera poses in Matterport3D.
Through experimentation, we found that treating the 12th camera pose, out of the 18 viewpoints (comprising 6 rotations and 3 tilt angles) in each panorama, identically to the second skybox image did not result in any issues.
We define the remaining poses for the \textit{skybox} images by rotating 90\textdegree\ in each direction from the second pose.
We adhere to the official benchmark split, utilizing 61 scenes for training, 11 for validation, and 18 for testing. 
For two-view pose estimation, it is necessary to create pairs of overlapped images. 
We achieve this by transforming ERP depth maps between frames within the spherical coordinate system.
Pixels where the depth difference is below a specified threshold, e.g. 0.1, are classified as inliers.
Subsequently, we compare the ratio of these inliers to the total number of pixels.
We organize both the training and testing datasets based on the overlap ratio of image pairs and the benchmark split. 
Specifically, images with the overlap ratio exceeding 30\% are distributed into respective training and testing splits. 
As a result, the training set contains 44,700 pairs, while the test set comprises 4,575 pairs. 
We resize the resolution of ERP images and depth maps to $640 \times 320$.
\vspace*{-1mm}
\paragraph{Stanford2D3D Dataset} Stanford2D3D \citep{armeni2017joint} consists of data scanned from six large-scale indoor spaces collected from three distinct buildings. 
This dataset contains a relatively small number of 1,413 panorama images and, therefore, is utilized exclusively for testing purposes.
We assess the overlap ratio between frames and include them in the test split if their ratio exceeds 50\%.
A total of 3,460 pairs are incorporated into the test set. 
During testing, we resize the resolution to $640 \times 320$.
\paragraph{EgoNeRF and OmniPhotos Dataset}
\vspace*{-1mm}
EgoNeRF \citep{choi2023balanced} introduces 11 synthetic scenes created with Blender \citep{blender} and 11 real scenes captured with a RICOH THETA V camera. OmniPhotos \citep{bertel2020omniphotos} provides a dataset captured with an Insta360 ONE X camera.
Both datasets contain egocentric scenes captured with a casually rotating camera stick.
Consequently, their rotation axes, pole regions, or camera height change, resulting in different distortions compared to Matterport3D or Stanford2D3D.
%
We present additional qualitative results from these datasets to validate our method.
%
\vspace*{-1mm}
\paragraph{Implementation Details} \label{section:experiment_detail}
We employ the AdamW \citep{loshchilov2017decoupled} optimizer with a weight-decay factor of $10^{-2}$, a learning rate of $5 \cdot 10^{-6}$ for multiscale feature extractor, and $10^{-4}$ for the SSAM and the Geodesic Flow Refiner.
EDM is trained for 300,000 steps with a batch size of 4 in a single RTX 3090 GPU, which takes approximately two days to complete. 
%
%
%
During evaluation, the balanced sampling approach using kernel density estimation \citep{edstedt2023dkm} tends to establish correspondences primarily in concentrated areas with high probability distributions, making it unsuitable for omnidirectional images.
%
%
Thus, we randomly sample up to 5,000 matches after certainty filtering with a threshold of 0.8 to ensure correspondences cover the entire area.
%
%

\begin{figure*}[t]
    \centering
    \includegraphics[width=0.95\linewidth]{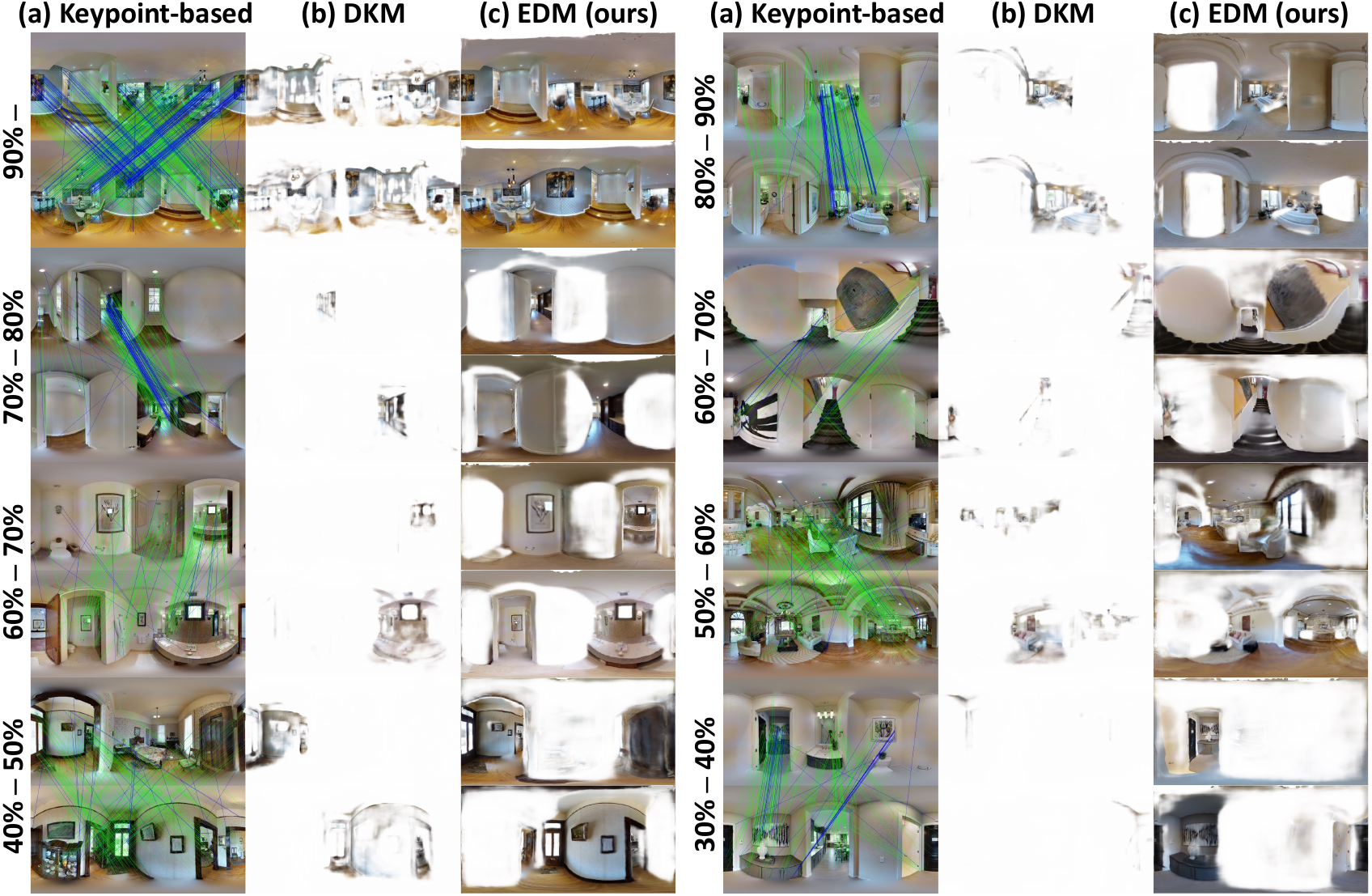}
    \caption{Qualitative results on Matterport3D. 
    (a) The blue lines represent the results of matching points from SPHORB \cite{zhao2015sphorb}; the green lines correspond to SphereGlue \cite{gava2023sphereglue}. Both (b) DKM \cite{edstedt2023dkm} and (c) EDM depict the outcomes of multiplying the warped image with the certainty map. 
    EDM can estimate dense and accurate matches even in the presence of distortions and severe occlusions.
    The numbers beside the images represent the overlap ratio, reflecting the difficulty of matching. Smaller numbers indicate more challenging scenes.
    }
    \label{fig:qualitative}
    \vspace*{-3mm}
\end{figure*}

\subsection{Experimental Results}
%
We compare our proposed method EDM with four different methods: 
1) SPHORB \citep{zhao2015sphorb} is a hand-crafted keypoint-based feature matching algorithm. 
2) SphereGlue \citep{gava2023sphereglue} is a learning-based keypoint matching method.
Both SPHORB \citep{zhao2015sphorb} and SphereGlue \citep{gava2023sphereglue} are specifically designed for spherical images. 
3) DKM \citep{edstedt2023dkm} and 4) RoMa \citep{edstedt2023roma} are state-of-the-art dense matching algorithms for perspective images.
To estimate the essential matrix and the relative pose for spherical cameras, \citet{solarte2021robust} proposed a normalization strategy and non-linear optimization within the classic 8-point algorithm.
We adopt this for two-view pose estimation in all quantitative comparisons.
%
\begin{table}[t]
    \centering
    \resizebox{0.5\textwidth}{!}{
    \begin{tabular}{l|c|c|ccc}
        \toprule
        \multirow{2}{*}{Method} &\multirow{2}{*}{Image} &\multirow{2}{*}{Feature} &\multicolumn{3}{c}{AUC} \\
        & & & @5\textdegree & @10\textdegree & @20\textdegree \\
        \midrule
        SPHORB \cite{zhao2015sphorb} &ERP &sparse &0.38 &1.41 &3.99 \\
        SphereGlue \cite{gava2023sphereglue} &ERP &sparse &11.29 &19.95 &31.10 \\
        \midrule
        DKM \cite{edstedt2023dkm} &persepctive &dense &18.43 &28.50 &38.44 \\
        RoMa \cite{edstedt2023roma} &perspective &dense &12.45 &22.37 &34.24 \\
        \midrule
        \textbf{EDM (ours)} &ERP &dense &\textbf{45.15} &\textbf{60.99} &\textbf{73.60} \\
        \bottomrule
    \end{tabular}
    }
    \caption{Quantitative comparison on Matterport3D with recent algorithms.  EDM improves AUC@5\textdegree\ by 26.72.}
    \label{table:mp3d}
\end{table}

\begin{table}[t]
    \centering
    \resizebox{0.5\textwidth}{!}{
    \begin{tabular}{l|c|c|ccc|}
        \toprule
        \multirow{2}{*}{Method} &\multirow{2}{*}{Image} &\multirow{2}{*}{Feature} &\multicolumn{3}{c|}{AUC} \\
        & & & @5\textdegree & @10\textdegree & @20\textdegree \\
        \midrule
        SPHORB \cite{zhao2015sphorb} &ERP &sparse &0.14 &1.01 &4.08 \\
        SphereGlue \cite{gava2023sphereglue} &ERP &sparse &11.25 &22.41 &36.57 \\
        \midrule
        DKM \cite{edstedt2023dkm} &perspective &dense &12.46 &22.18 &34.13 \\
        RoMa \cite{edstedt2023roma} &perspective &dense &11.48 &22.52 &37.07 \\
        \midrule
        \textbf{EDM (ours)} &ERP &dense &\bf{55.08} &\bf{71.65} &\bf{82.72} \\
        \bottomrule
    \end{tabular}
    }
    \caption{Quantitative comparison on Stanford2D3D with recent algorithms.  EDM improve AUC@5\textdegree\ by 42.62.}
    \label{table:stfd}
    
\end{table}

\begin{figure}
    \centering
    \includegraphics[width=0.95\linewidth]{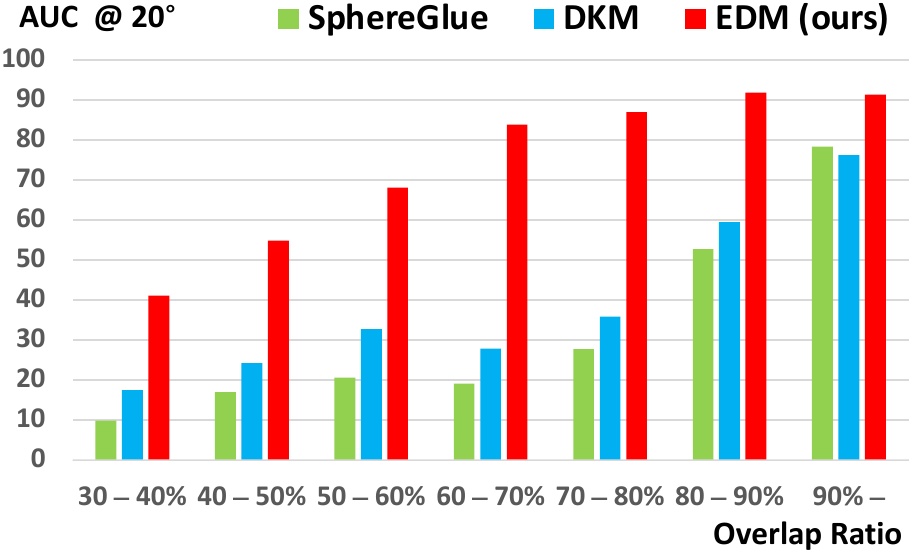}
    \caption{Performance with respect to the overlap ratio.
    This highlights the robustness of EDM in scenarios with varying levels of overlap, particularly in challenging conditions where the overlap ratio is limited.
    }
    \label{fig:performance}
\end{figure}
Table \ref{table:mp3d} shows the quantitative results of the pose estimation in Matterport3D. 
Despite SPHORB and SphereGlue being designed for the ERP images, the presence of textureless or repetitive regions, which are common in indoor environments of Matterport3D, leads to performance degradation in the keypoint-based methods.
SPHORB fails to estimate the essential matrix correctly due to the limited number of matching points.
EDM demonstrates significantly higher performance than all the other methods.

Figure \ref{fig:qualitative} illustrates the qualitative results in Matterport3D. 
%
The previous methods designed for perspective images, such as DKM and RoMa, exhibit good matching ability but encounter challenges when confronted with the distortions of ERP.
While SphereGlue and SPHORB perform well in discriminative regions, their performance deteriorates as the overlap ratio decreases, resulting in numerous false positive matches.
In contrast, EDM can estimate dense correspondences regardless of occlusion and textureless areas.
Due to the similarity in results between DKM and RoMa, we have only included the former to maintain a concise visualization.
%
Experimental results in Fig.\ref{fig:performance} depict the relationship between image overlap ratio and AUC@20\textdegree\ performance. 
As expected, a decrease in the overlap ratio leads to severe performance degradation in the previous works.
On the other hand, our proposed method demonstrates robustness in more challenging scenes, maintaining similar performance levels until the overlap decreases to 60\%, compared to other methods.
%



\begin{figure}
    \centering
    \includegraphics[width=1.0\linewidth]{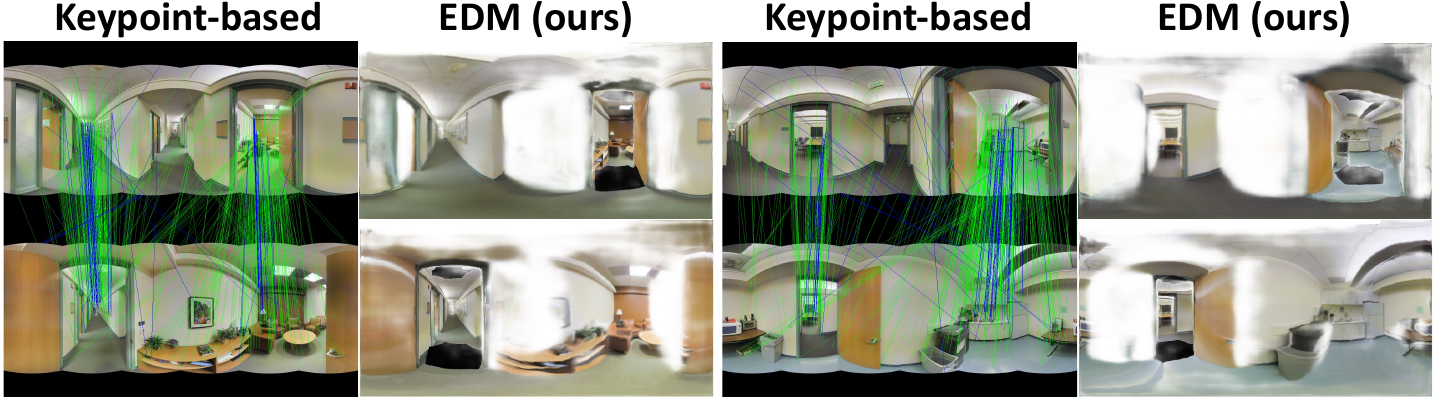}
    \captionof{figure}{Qualitative results on Stanford2D3D. The blue and green lines correspond to SPHORB and SphereGlue. }
    \label{fig:stfd}
\end{figure}
%
%
For a fair comparison, we use another benchmark dataset, Stanford2D3D. 
We validate EDM using a model trained on Matterport3D without additional training on Stanford2D3D.
%
%
In Table \ref{table:stfd}, EDM outperforms the previous works by a significant margin, especially in scenes with severe occlusion.
The certainty map demonstrates EDM's robustness, particularly in handling occluded scenes.
Additionally, although the panorama images in Stanford2D3D contain missing regions in the upper and lower parts of the sphere, the proposed spherical positional embedding enables the network to predict matching correspondences accurately, as shown in Fig. \ref{fig:stfd}.
%

\subsection{Additional Qualitative Results}
\begin{figure}[t]
    \centering
    \includegraphics[width=1.0\linewidth]{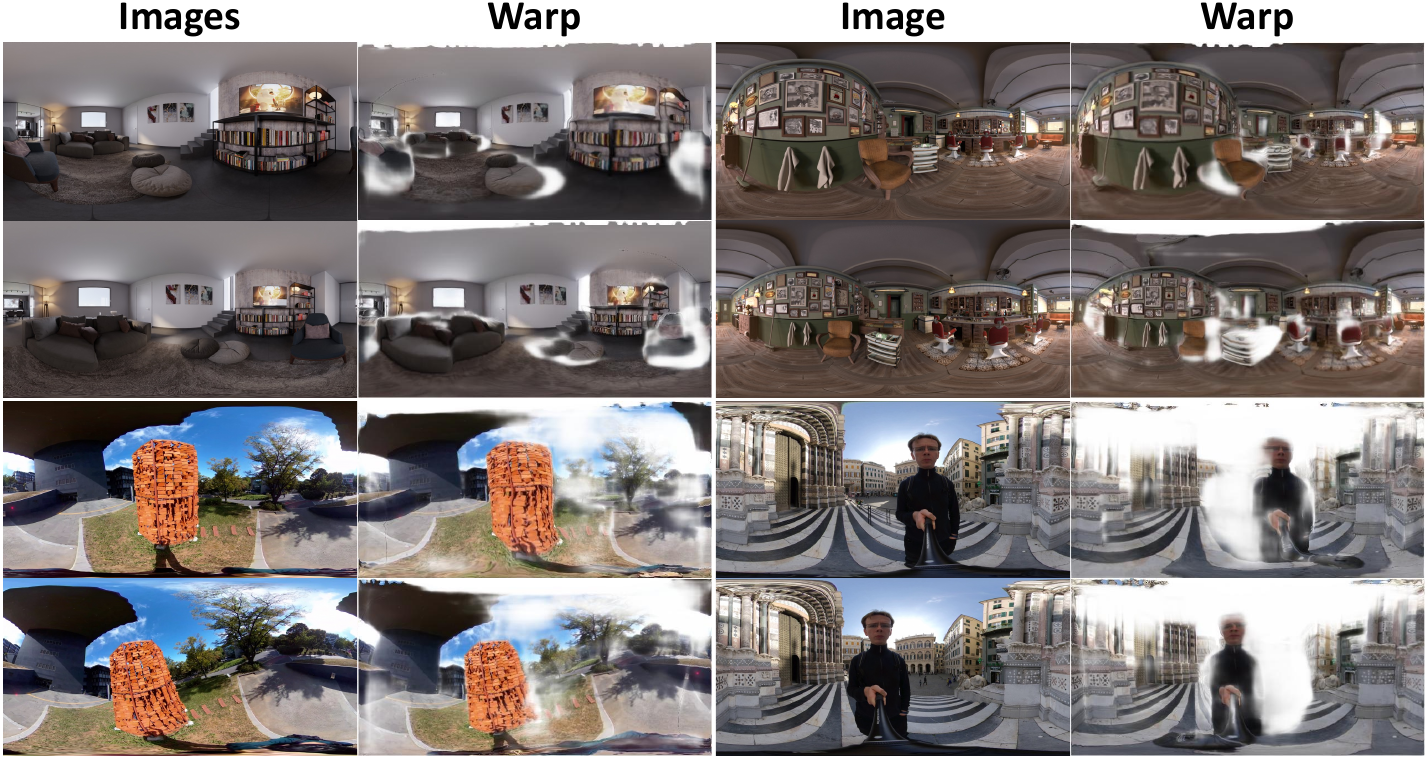}
    \caption{Qualitative results on EgoNeRF \cite{choi2023balanced} and OmniPhotos \cite{bertel2020omniphotos}.
    Despite being primarily trained on indoor scenes, EDM effectively estimates dense matching on these datasets, demonstrating its generalization capability across diverse environments.}
    \label{fig:additional}
\end{figure}
To demonstrate the robust performance of our method across diverse environments, we qualitatively validate EDM using additional datasets such as EgoNeRF and OmniPhotos.
As it is primarily trained on indoor environments \citep{chang2017matterport3d} where the camera is oriented parallel to gravity, severely slanted image pairs of rotational scenes or outdoor environments may cause EDM to fail in accurately estimating correspondences.
However, despite these differences in settings, EDM demonstrates the ability to conduct dense feature matching robustly, as shown in Fig. \ref{fig:additional}.

\begin{figure}[h]
    \centering
    \includegraphics[width=0.9\linewidth]{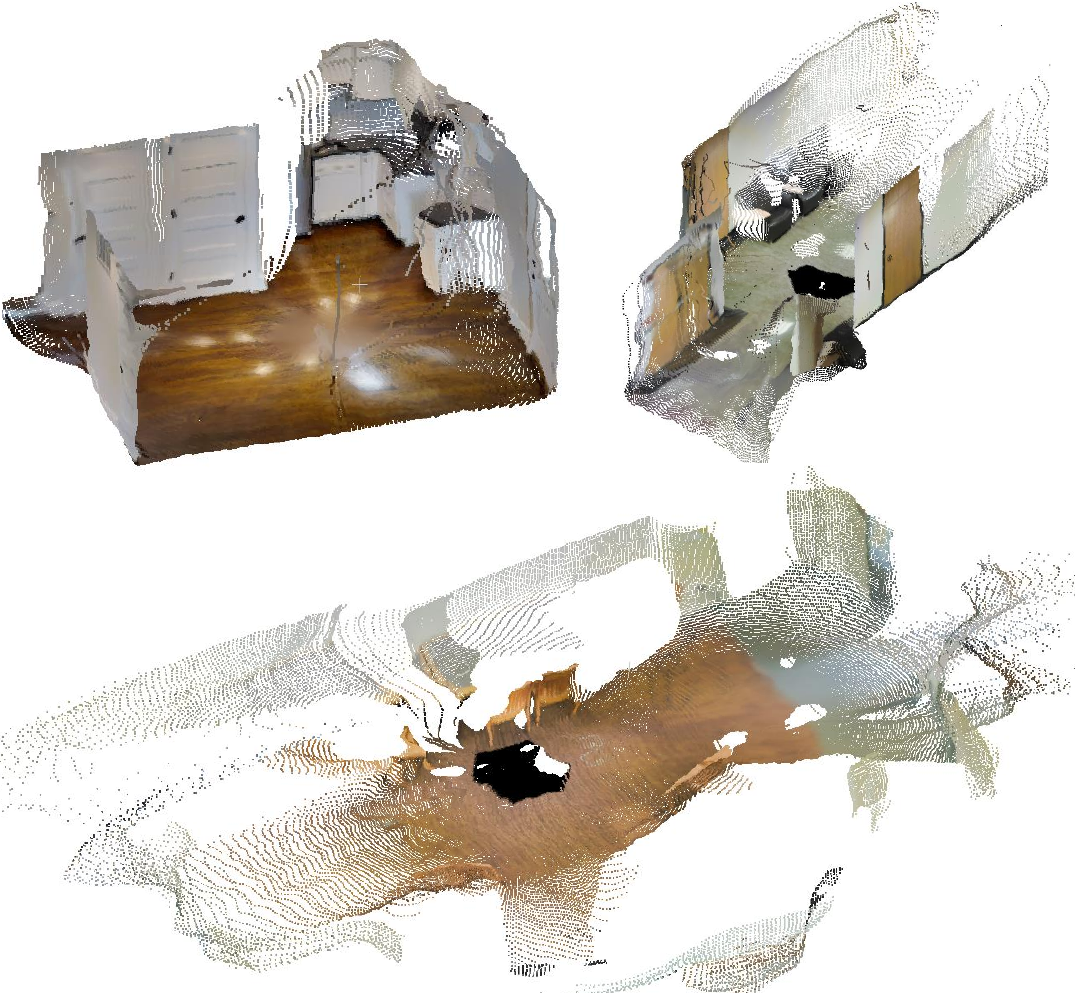}
    \caption{Triangulation results on Matterport3D and Stanford2D3D. These point clouds are generated through spherical triangulation using the estimated poses between two omnidirectional images. Our method can reconstruct dense point clouds in textureless regions, which are particularly challenging in indoor environments.}
    \label{fig:3dpoints}
\end{figure}

Furthermore, we demonstrate the applicability of our method to various omnidirectional downstream tasks. As shown in Fig. \ref{fig:3dpoints}, our approach successfully performs triangulation from pairs of omnidirectional images. By leveraging EDM's capability to predict dense correspondences, the triangulated points yield a dense 3D reconstruction. For a more comprehensive discussion, please refer to the supplementary materials.

\subsection{Ablation Study}
%
%
DKM's dependence on the pinhole camera model makes it inherently unsuitable for learning with ERP images.
%
%
To ensure the fair comparison, we modified the warping process in the loss function of DKM to support spherical cameras, resulting in DKM$^{*}$.
As shown in Table \ref{table:ablation}, this demonstrates the structural effectiveness of our proposed bidirectional coordinate transformation.
%
The proposed positional embeddings result in improvements based on the coordinate system of the spherical camera model.
We observe that utilizing a 3D grid input of Cartesian coordinates yields better performance than 2D spherical ones.
Additionally, in our method, positional embedding with a linear layer slightly outperforms spherical positional encoding with sinusoidal \citep{li2023panoramic}.
Table \ref{table:ablation} also confirms the advantage of our rotational augmentation. Through this augmentation technique, we can effectively address the challenge of a limited number of datasets for omnidirectional images in dense matching tasks.


\begin{table}
    \centering
    \resizebox{1.0\linewidth}{!}{
    \begin{tabular}{l|ccc|ccc}
        \toprule
        \multirow{2}{*}{Method} & \multirow{1}{*}{Positional} & \multirow{1}{*}{Bidirectional} & \multirow{1}{*}{Rotational} &\multicolumn{3}{c}{AUC} \\
        & Embedding & Transformation & Augmentation & @5\textdegree & @10\textdegree & @20\textdegree \\
        \midrule
        DKM$^{*}$ & 2D linear & - & - & 19.83 & 33.06 & 46.24 \\
        Ours & 2D linear & \checkmark & - & 29.67 & 45.90 & 60.82 \\
        Ours & 2D linear & \checkmark & \checkmark & 35.03 & 51.14 & 65.07 \\
        Ours & 3D linear & \checkmark & - &34.64 &50.82 &65.16 \\
        \textbf{Ours} & \textbf{3D linear} & \checkmark & \checkmark & \textbf{45.15} & \textbf{60.99} & \textbf{73.60} \\
        Ours & 3D sinusoidal & \checkmark & \checkmark & 42.39 & 58.27 & 70.98 \\
        \bottomrule
    \end{tabular}
    }
    \caption{\label{table:ablation}Ablation study for the proposed method. DKM$^{*}$ indicates the DKM model trained on Matterport3D with a modified loss function for ERP images. Compared to DKM$^{*}$, our method enhances performance through the proposed spherical positional embedding in SSAM, bidirectional transformation via Geodesic Flow Refinement, and rotational augmentation.}
    \label{table:ablation_aug}
\end{table}

\section{Conclusion, Limitations, and Future Work}
\label{section:conclusion}
In this paper, we present, for the first time, a novel dense feature matching method tailored for omnidirectional images. 
Leveraging the foundational principles of DKM, we integrate the inherent characteristics of the spherical camera model into our dense matching process using geodesic flow fields. 
This integration instills distortion awareness within the network, thereby enhancing its performance specifically for ERP images. 
However, it is important to note that our method is predominantly trained on indoor datasets where the camera is vertically oriented, rendering it somewhat vulnerable to extreme rotations or outdoor environments. 
To address this limitation, future endeavors will focus on diversifying the training data and data augmentation to encompass a wider range of environments, fortifying the robustness of our network. 
Furthermore, we aim to extend our method into downstream tasks, particularly for visual localization and mapping applications for omnidirectional images.

{
    \small
    \bibliographystyle{ieeenat_fullname}
    \bibliography{main}

\begin{thebibliography}{71}
\providecommand{\natexlab}[1]{#1}
\providecommand{\url}[1]{\texttt{#1}}
\expandafter\ifx\csname urlstyle\endcsname\relax
  \providecommand{\doi}[1]{doi: #1}\else
  \providecommand{\doi}{doi: \begingroup \urlstyle{rm}\Url}\fi

\bibitem[Amalia and Fitriyansah(2023)]{amalia2023case}
Friska Amalia and Ahmad Fitriyansah.
\newblock Case study of 360 image viewer software utilization in interior design presentation to improve product immersion.
\newblock In \emph{ICCED}. IEEE, 2023.

\bibitem[Armeni et~al.(2017)Armeni, Sax, Zamir, and Savarese]{armeni2017joint}
Iro Armeni, Sasha Sax, Amir~R Zamir, and Silvio Savarese.
\newblock Joint 2d-3d-semantic data for indoor scene understanding.
\newblock \emph{arXiv preprint arXiv:1702.01105}, 2017.

\bibitem[Bergmann et~al.(2021)Bergmann, Pinto, da~Silveira, and Jung]{bergmann2021gravity}
Matheus~A Bergmann, Paulo~GL Pinto, Thiago~LT da Silveira, and Cl{\'a}udio~R Jung.
\newblock Gravity alignment for single panorama depth inference.
\newblock In \emph{SIBGRAPI}. IEEE, 2021.

\bibitem[Bertel et~al.(2020)Bertel, Yuan, Lindroos, and Richardt]{bertel2020omniphotos}
Tobias Bertel, Mingze Yuan, Reuben Lindroos, and Christian Richardt.
\newblock Omniphotos: casual 360 vr photography.
\newblock \emph{ACM Transactions on Graphics (TOG)}, 39\penalty0 (6):\penalty0 1--12, 2020.

\bibitem[Chang et~al.(2017)Chang, Dai, Funkhouser, Halber, Niessner, Savva, Song, Zeng, and Zhang]{chang2017matterport3d}
Angel Chang, Angela Dai, Thomas Funkhouser, Maciej Halber, Matthias Niessner, Manolis Savva, Shuran Song, Andy Zeng, and Yinda Zhang.
\newblock Matterport3d: Learning from rgb-d data in indoor environments.
\newblock \emph{arXiv preprint arXiv:1709.06158}, 2017.

\bibitem[Chen et~al.(2022)Chen, Wang, and Liu]{chen2022text2light}
Zhaoxi Chen, Guangcong Wang, and Ziwei Liu.
\newblock Text2light: Zero-shot text-driven hdr panorama generation.
\newblock \emph{ACM Transactions on Graphics (TOG)}, 41\penalty0 (6):\penalty0 1--16, 2022.

\bibitem[Choi et~al.(2023)Choi, Kim, and Kim]{choi2023balanced}
Changwoon Choi, Sang~Min Kim, and Young~Min Kim.
\newblock Balanced spherical grid for egocentric view synthesis.
\newblock In \emph{CVPR}, 2023.

\bibitem[Choi et~al.(2024)Choi, Jang, and Kim]{choi2024omnilocalrf}
Dongyoung Choi, Hyeonjoong Jang, and Min~H Kim.
\newblock Omnilocalrf: Omnidirectional local radiance fields from dynamic videos.
\newblock \emph{arXiv preprint arXiv:2404.00676}, 2024.

\bibitem[Cohen et~al.(2018)Cohen, Geiger, K{\"o}hler, and Welling]{cohen2018spherical}
Taco~S Cohen, Mario Geiger, Jonas K{\"o}hler, and Max Welling.
\newblock Spherical cnns.
\newblock \emph{arXiv preprint arXiv:1801.10130}, 2018.

\bibitem[Community(2018)]{blender}
Blender~Online Community.
\newblock \emph{Blender - a 3D modelling and rendering package}.
\newblock Blender Foundation, Stichting Blender Foundation, Amsterdam, 2018.

\bibitem[Coors et~al.(2018)Coors, Condurache, and Geiger]{coors2018spherenet}
Benjamin Coors, Alexandru~Paul Condurache, and Andreas Geiger.
\newblock Spherenet: Learning spherical representations for detection and classification in omnidirectional images.
\newblock In \emph{ECCV}, 2018.

\bibitem[da~Silveira et~al.(2022)da~Silveira, Pinto, Murrugarra-Llerena, and Jung]{da20223d}
Thiago~LT da Silveira, Paulo~GL Pinto, Jeffri Murrugarra-Llerena, and Cl{\'a}udio~R Jung.
\newblock 3d scene geometry estimation from 360 imagery: A survey.
\newblock \emph{ACM Computing Surveys}, 55\penalty0 (4):\penalty0 1--39, 2022.

\bibitem[DeTone et~al.(2018)DeTone, Malisiewicz, and Rabinovich]{detone2018superpoint}
Daniel DeTone, Tomasz Malisiewicz, and Andrew Rabinovich.
\newblock Superpoint: Self-supervised interest point detection and description.
\newblock In \emph{CVPR Workshops}, 2018.

\bibitem[Eder et~al.(2020)Eder, Shvets, Lim, and Frahm]{eder2020tangent}
Marc Eder, Mykhailo Shvets, John Lim, and Jan-Michael Frahm.
\newblock Tangent images for mitigating spherical distortion.
\newblock In \emph{CVPR}, 2020.

\bibitem[Edstedt et~al.(2023{\natexlab{a}})Edstedt, Athanasiadis, Wadenb{\"a}ck, and Felsberg]{edstedt2023dkm}
Johan Edstedt, Ioannis Athanasiadis, M{\aa}rten Wadenb{\"a}ck, and Michael Felsberg.
\newblock D{K}{M}: Dense kernelized feature matching for geometry estimation.
\newblock In \emph{CVPR}, 2023{\natexlab{a}}.

\bibitem[Edstedt et~al.(2023{\natexlab{b}})Edstedt, Sun, B{\"o}kman, Wadenb{\"a}ck, and Felsberg]{edstedt2023roma}
Johan Edstedt, Qiyu Sun, Georg B{\"o}kman, M{\aa}rten Wadenb{\"a}ck, and Michael Felsberg.
\newblock Roma: Revisiting robust losses for dense feature matching.
\newblock \emph{arXiv preprint arXiv:2305.15404}, 2023{\natexlab{b}}.

\bibitem[Esteves et~al.(2018)Esteves, Allen-Blanchette, Makadia, and Daniilidis]{esteves2018learning}
Carlos Esteves, Christine Allen-Blanchette, Ameesh Makadia, and Kostas Daniilidis.
\newblock Learning so (3) equivariant representations with spherical cnns.
\newblock In \emph{ECCV}, 2018.

\bibitem[Ferm{\"u}ller and Aloimonos(2001)]{fermuller2001geometry}
Cornelia Ferm{\"u}ller and Yiannis Aloimonos.
\newblock Geometry of eye design: Biology and technology.
\newblock In \emph{Multi-Image Analysis: 10th International Workshop on Theoretical Foundations of Computer Vision Dagstuhl Castle, Germany, March 12--17, 2000 Revised Papers}, pages 22--38. Springer, 2001.

\bibitem[Gava et~al.(2023)Gava, Mukunda, Habtegebrial, Raue, Palacio, and Dengel]{gava2023sphereglue}
Christiano Gava, Vishal Mukunda, Tewodros Habtegebrial, Federico Raue, Sebastian Palacio, and Andreas Dengel.
\newblock Sphereglue: Learning keypoint matching on high resolution spherical images.
\newblock In \emph{CVPR Workshops}, 2023.

\bibitem[Gava et~al.(2024)Gava, Cho, Raue, Palacio, Pagani, and Dengel]{gava2024spherecraft}
Christiano Gava, Yunmin Cho, Federico Raue, Sebastian Palacio, Alain Pagani, and Andreas Dengel.
\newblock Spherecraft: A dataset for spherical keypoint detection, matching and camera pose estimation.
\newblock In \emph{WACV}, 2024.

\bibitem[Guerrero-Viu et~al.(2020)Guerrero-Viu, Fernandez-Labrador, Demonceaux, and Guerrero]{guerrero2020s}
Julia Guerrero-Viu, Clara Fernandez-Labrador, C{\'e}dric Demonceaux, and Jose~J Guerrero.
\newblock What’s in my room? object recognition on indoor panoramic images.
\newblock In \emph{ICRA}. IEEE, 2020.

\bibitem[He et~al.(2016)He, Zhang, Ren, and Sun]{he2016deep}
Kaiming He, Xiangyu Zhang, Shaoqing Ren, and Jian Sun.
\newblock Deep residual learning for image recognition.
\newblock In \emph{CVPR}, 2016.

\bibitem[Hutchcroft et~al.(2022)Hutchcroft, Li, Boyadzhiev, Wan, Wang, and Kang]{hutchcroft2022covispose}
Will Hutchcroft, Yuguang Li, Ivaylo Boyadzhiev, Zhiqiang Wan, Haiyan Wang, and Sing~Bing Kang.
\newblock Covispose: Co-visibility pose transformer for wide-baseline relative pose estimation in 360\textdegree \ indoor panoramas.
\newblock In \emph{ECCV}. Springer, 2022.

\bibitem[Jiang et~al.(2019)Jiang, Huang, Kashinath, Marcus, Niessner, et~al.]{jiang2019spherical}
Chiyu Jiang, Jingwei Huang, Karthik Kashinath, Philip Marcus, Matthias Niessner, et~al.
\newblock Spherical cnns on unstructured grids.
\newblock \emph{arXiv preprint arXiv:1901.02039}, 2019.

\bibitem[Jiang et~al.(2021)Jiang, Sheng, Zhu, Dong, and Huang]{jiang2021unifuse}
Hualie Jiang, Zhe Sheng, Siyu Zhu, Zilong Dong, and Rui Huang.
\newblock Unifuse: Unidirectional fusion for 360 panorama depth estimation.
\newblock \emph{IEEE Robotics and Automation Letters}, 6\penalty0 (2):\penalty0 1519--1526, 2021.

\bibitem[Kim et~al.(2024)Kim, Meuleman, Jang, Tompkin, and Kim]{kim2024omnisdf}
Hakyeong Kim, Andreas Meuleman, Hyeonjoong Jang, James Tompkin, and Min~H Kim.
\newblock Omnisdf: Scene reconstruction using omnidirectional signed distance functions and adaptive binoctrees.
\newblock \emph{arXiv preprint arXiv:2404.00678}, 2024.

\bibitem[Lee et~al.(2000)Lee, You, and Neumann]{lee2000large}
Jong~Weon Lee, Suya You, and Ulrich Neumann.
\newblock Large motion estimation for omnidirectional vision.
\newblock In \emph{Proceedings IEEE Workshop on Omnidirectional Vision (Cat. No. PR00704)}, pages 161--168. IEEE, 2000.

\bibitem[Li et~al.(2022{\natexlab{a}})Li, Wang, Liu, Ran, Xu, and Guo]{li2022decoupling}
Kunhong Li, Longguang Wang, Li Liu, Qing Ran, Kai Xu, and Yulan Guo.
\newblock Decoupling makes weakly supervised local feature better.
\newblock In \emph{CVPR}, 2022{\natexlab{a}}.

\bibitem[Li et~al.(2024)Li, Huang, Yeung, and Cheng]{li2024omnigs}
Longwei Li, Huajian Huang, Sai-Kit Yeung, and Hui Cheng.
\newblock Omnigs: Omnidirectional gaussian splatting for fast radiance field reconstruction using omnidirectional images.
\newblock \emph{arXiv preprint arXiv:2404.03202}, 2024.

\bibitem[Li et~al.(2023{\natexlab{a}})Li, Wang, Yuan, Shen, Sheng, and Dong]{li2023mathcal}
Meng Li, Senbo Wang, Weihao Yuan, Weichao Shen, Zhe Sheng, and Zilong Dong.
\newblock S2{N}et: Accurate panorama depth estimation on spherical surface.
\newblock \emph{IEEE Robotics and Automation Letters}, 8\penalty0 (2):\penalty0 1053--1060, 2023{\natexlab{a}}.

\bibitem[Li et~al.(2023{\natexlab{b}})Li, Cao, Zhao, Li, Zhang, and Raj]{li2023panoramic}
Xiang Li, Haoyuan Cao, Shijie Zhao, Junlin Li, Li Zhang, and Bhiksha Raj.
\newblock Panoramic video salient object detection with ambisonic audio guidance.
\newblock In \emph{AAAI}, 2023{\natexlab{b}}.

\bibitem[Li et~al.(2021)Li, Yan, Duan, and Ren]{li2021panodepth}
Yuyan Li, Zhixin Yan, Ye Duan, and Liu Ren.
\newblock Panodepth: A two-stage approach for monocular omnidirectional depth estimation.
\newblock In \emph{3DV}. IEEE, 2021.

\bibitem[Li et~al.(2022{\natexlab{b}})Li, Guo, Yan, Huang, Duan, and Ren]{li2022omnifusion}
Yuyan Li, Yuliang Guo, Zhixin Yan, Xinyu Huang, Ye Duan, and Liu Ren.
\newblock Omnifusion: 360 monocular depth estimation via geometry-aware fusion.
\newblock In \emph{CVPR}, 2022{\natexlab{b}}.

\bibitem[Liu et~al.(2019)Liu, Shen, Lin, Peng, Bao, and Zhou]{liu2019gift}
Yuan Liu, Zehong Shen, Zhixuan Lin, Sida Peng, Hujun Bao, and Xiaowei Zhou.
\newblock Gift: Learning transformation-invariant dense visual descriptors via group cnns.
\newblock \emph{Advances in Neural Information Processing Systems}, 32, 2019.

\bibitem[Loshchilov and Hutter(2017)]{loshchilov2017decoupled}
Ilya Loshchilov and Frank Hutter.
\newblock Decoupled weight decay regularization.
\newblock \emph{arXiv preprint arXiv:1711.05101}, 2017.

\bibitem[Lowe(2004)]{lowe2004distinctive}
David~G Lowe.
\newblock Distinctive image features from scale-invariant keypoints.
\newblock \emph{International journal of computer vision}, 60:\penalty0 91--110, 2004.

\bibitem[Ma et~al.(2024)Ma, Zhan, and Jin]{ma2024fastscene}
Yikun Ma, Dandan Zhan, and Zhi Jin.
\newblock Fastscene: Text-driven fast 3d indoor scene generation via panoramic gaussian splatting.
\newblock \emph{arXiv preprint arXiv:2405.05768}, 2024.

\bibitem[Matzen et~al.(2017)Matzen, Cohen, Evans, Kopf, and Szeliski]{matzen2017low}
Kevin Matzen, Michael~F Cohen, Bryce Evans, Johannes Kopf, and Richard Szeliski.
\newblock Low-cost 360 stereo photography and video capture.
\newblock \emph{ACM Transactions on Graphics (TOG)}, 36\penalty0 (4):\penalty0 1--12, 2017.

\bibitem[Melekhov et~al.(2019)Melekhov, Tiulpin, Sattler, Pollefeys, Rahtu, and Kannala]{melekhov2019dgc}
Iaroslav Melekhov, Aleksei Tiulpin, Torsten Sattler, Marc Pollefeys, Esa Rahtu, and Juho Kannala.
\newblock Dgc-net: Dense geometric correspondence network.
\newblock In \emph{WACV}. IEEE, 2019.

\bibitem[Menegatti et~al.(2004)Menegatti, Maeda, and Ishiguro]{menegatti2004image}
Emanuele Menegatti, Takeshi Maeda, and Hiroshi Ishiguro.
\newblock Image-based memory for robot navigation using properties of omnidirectional images.
\newblock \emph{Robotics and Autonomous Systems}, 47\penalty0 (4):\penalty0 251--267, 2004.

\bibitem[Nejatishahidin et~al.(2023)Nejatishahidin, Hutchcroft, Narayana, Boyadzhiev, Li, Khosravan, Ko{\v{s}}eck{\'a}, and Kang]{nejatishahidin2023graph}
Negar Nejatishahidin, Will Hutchcroft, Manjunath Narayana, Ivaylo Boyadzhiev, Yuguang Li, Naji Khosravan, Jana Ko{\v{s}}eck{\'a}, and Sing~Bing Kang.
\newblock Graph-covis: Gnn-based multi-view panorama global pose estimation.
\newblock In \emph{CVPR}, 2023.

\bibitem[Nelson and Aloimonos(1988)]{nelson1988finding}
Randal~C Nelson and John Aloimonos.
\newblock Finding motion parameters from spherical motion fields (or the advantages of having eyes in the back of your head).
\newblock \emph{Biological cybernetics}, 58\penalty0 (4):\penalty0 261--273, 1988.

\bibitem[Pandey et~al.(2011)Pandey, McBride, and Eustice]{pandey2011ford}
Gaurav Pandey, James~R McBride, and Ryan~M Eustice.
\newblock Ford campus vision and lidar data set.
\newblock \emph{The International Journal of Robotics Research}, 30\penalty0 (13):\penalty0 1543--1552, 2011.

\bibitem[Revaud et~al.(2019)Revaud, De~Souza, Humenberger, and Weinzaepfel]{revaud2019r2d2}
Jerome Revaud, Cesar De~Souza, Martin Humenberger, and Philippe Weinzaepfel.
\newblock R2d2: Reliable and repeatable detector and descriptor.
\newblock \emph{Advances in neural information processing systems}, 32, 2019.

\bibitem[Rey-Area et~al.(2022)Rey-Area, Yuan, and Richardt]{rey2022360monodepth}
Manuel Rey-Area, Mingze Yuan, and Christian Richardt.
\newblock 360monodepth: High-resolution 360deg monocular depth estimation.
\newblock In \emph{CVPR}, 2022.

\bibitem[Rublee et~al.(2011)Rublee, Rabaud, Konolige, and Bradski]{rublee2011orb}
Ethan Rublee, Vincent Rabaud, Kurt Konolige, and Gary Bradski.
\newblock Orb: An efficient alternative to sift or surf.
\newblock In \emph{ICCV}. Ieee, 2011.

\bibitem[Sarlin et~al.(2020)Sarlin, DeTone, Malisiewicz, and Rabinovich]{sarlin2020superglue}
Paul-Edouard Sarlin, Daniel DeTone, Tomasz Malisiewicz, and Andrew Rabinovich.
\newblock Superglue: Learning feature matching with graph neural networks.
\newblock In \emph{CVPR}, pages 4938--4947, 2020.

\bibitem[Saurer et~al.(2010)Saurer, Fraundorfer, and Pollefeys]{saurer2010omnitour}
Olivier Saurer, Friedrich Fraundorfer, and Marc Pollefeys.
\newblock Omnitour: Semi-automatic generation of interactive virtual tours from omnidirectional video.
\newblock In \emph{3DPVT}, 2010.

\bibitem[Schonberger and Frahm(2016)]{schonberger2016structure}
Johannes~L Schonberger and Jan-Michael Frahm.
\newblock Structure-from-motion revisited.
\newblock In \emph{CVPR}, 2016.

\bibitem[Shen et~al.(2022)Shen, Lin, Liao, Nie, Zheng, and Zhao]{shen2022panoformer}
Zhijie Shen, Chunyu Lin, Kang Liao, Lang Nie, Zishuo Zheng, and Yao Zhao.
\newblock Panoformer: Panorama transformer for indoor 360\textdegree \ depth estimation.
\newblock In \emph{ECCV}. Springer, 2022.

\bibitem[Snippe and Koenderink(1992)]{snippe1992discrimination}
Herman~P Snippe and Jan~J Koenderink.
\newblock Discrimination thresholds for channel-coded systems.
\newblock \emph{Biological cybernetics}, 66\penalty0 (6):\penalty0 543--551, 1992.

\bibitem[Solarte et~al.(2021)Solarte, Wu, Lu, Tsai, Chiu, and Sun]{solarte2021robust}
Bolivar Solarte, Chin-Hsuan Wu, Kuan-Wei Lu, Yi-Hsuan Tsai, Wei-Chen Chiu, and Min Sun.
\newblock Robust 360-8pa: Redesigning the normalized 8-point algorithm for 360-fov images.
\newblock In \emph{ICRA}. IEEE, 2021.

\bibitem[Su and Grauman(2017)]{su2017learning}
Yu-Chuan Su and Kristen Grauman.
\newblock Learning spherical convolution for fast features from 360 imagery.
\newblock \emph{Advances in neural information processing systems}, 30, 2017.

\bibitem[Sun et~al.(2021{\natexlab{a}})Sun, Sun, and Chen]{sun2021hohonet}
Cheng Sun, Min Sun, and Hwann-Tzong Chen.
\newblock Hohonet: 360 indoor holistic understanding with latent horizontal features.
\newblock In \emph{CVPR}, 2021{\natexlab{a}}.

\bibitem[Sun et~al.(2021{\natexlab{b}})Sun, Shen, Wang, Bao, and Zhou]{sun2021loftr}
Jiaming Sun, Zehong Shen, Yuang Wang, Hujun Bao, and Xiaowei Zhou.
\newblock Loftr: Detector-free local feature matching with transformers.
\newblock In \emph{CVPR}, 2021{\natexlab{b}}.

\bibitem[Tan et~al.(2022)Tan, Liu, Chen, Chen, Zhang, Shen, Ding, and Ji]{tan2022eco}
Dongli Tan, Jiang-Jiang Liu, Xingyu Chen, Chao Chen, Ruixin Zhang, Yunhang Shen, Shouhong Ding, and Rongrong Ji.
\newblock Eco-tr: Efficient correspondences finding via coarse-to-fine refinement.
\newblock In \emph{ECCV}. Springer, 2022.

\bibitem[Truong et~al.(2020)Truong, Danelljan, and Timofte]{truong2020glu}
Prune Truong, Martin Danelljan, and Radu Timofte.
\newblock Glu-net: Global-local universal network for dense flow and correspondences.
\newblock In \emph{CVPR}, 2020.

\bibitem[Truong et~al.(2021)Truong, Danelljan, Van~Gool, and Timofte]{truong2021learning}
Prune Truong, Martin Danelljan, Luc Van~Gool, and Radu Timofte.
\newblock Learning accurate dense correspondences and when to trust them.
\newblock In \emph{CVPR}, 2021.

\bibitem[Tyszkiewicz et~al.(2020)Tyszkiewicz, Fua, and Trulls]{tyszkiewicz2020disk}
Micha{\l} Tyszkiewicz, Pascal Fua, and Eduard Trulls.
\newblock Disk: Learning local features with policy gradient.
\newblock \emph{Advances in Neural Information Processing Systems}, 33:\penalty0 14254--14265, 2020.

\bibitem[Wang et~al.(2020)Wang, Yeh, Sun, Chiu, and Tsai]{wang2020bifuse}
Fu-En Wang, Yu-Hsuan Yeh, Min Sun, Wei-Chen Chiu, and Yi-Hsuan Tsai.
\newblock Bifuse: Monocular 360 depth estimation via bi-projection fusion.
\newblock In \emph{CVPR}, 2020.

\bibitem[Winters et~al.(2000)Winters, Gaspar, Lacey, and Santos-Victor]{winters2000omni}
Niall Winters, Jos{\'e} Gaspar, Gerard Lacey, and Jos{\'e} Santos-Victor.
\newblock Omni-directional vision for robot navigation.
\newblock In \emph{Proceedings IEEE Workshop on Omnidirectional Vision (Cat. No. PR00704)}, pages 21--28. IEEE, 2000.

\bibitem[Won et~al.(2020)Won, Seok, Cui, Pollefeys, and Lim]{won2020omnislam}
Changhee Won, Hochang Seok, Zhaopeng Cui, Marc Pollefeys, and Jongwoo Lim.
\newblock Omnislam: Omnidirectional localization and dense mapping for wide-baseline multi-camera systems.
\newblock In \emph{ICRA}. IEEE, 2020.

\bibitem[Xu et~al.(2020)Xu, Li, Zhang, and Le~Callet]{xu2020state}
Mai Xu, Chen Li, Shanyi Zhang, and Patrick Le~Callet.
\newblock State-of-the-art in 360 video/image processing: Perception, assessment and compression.
\newblock \emph{IEEE Journal of Selected Topics in Signal Processing}, 14\penalty0 (1):\penalty0 5--26, 2020.

\bibitem[Yu et~al.(2018)Yu, Wang, Peng, Gao, Yu, and Sang]{yu2018learning}
Changqian Yu, Jingbo Wang, Chao Peng, Changxin Gao, Gang Yu, and Nong Sang.
\newblock Learning a discriminative feature network for semantic segmentation.
\newblock In \emph{CVPR}, 2018.

\bibitem[Yun et~al.(2022)Yun, Lee, and Rhee]{yun2022improving}
Ilwi Yun, Hyuk-Jae Lee, and Chae~Eun Rhee.
\newblock Improving 360 monocular depth estimation via non-local dense prediction transformer and joint supervised and self-supervised learning.
\newblock In \emph{AAAI}, 2022.

\bibitem[Zhang et~al.(2019)Zhang, Liwicki, Smith, and Cipolla]{zhang2019orientation}
Chao Zhang, Stephan Liwicki, William Smith, and Roberto Cipolla.
\newblock Orientation-aware semantic segmentation on icosahedron spheres.
\newblock In \emph{ICCV}, 2019.

\bibitem[Zhang et~al.(2023{\natexlab{a}})Zhang, Zhao, Zhang, and Zollmann]{zhang2023survey}
Fanglue Zhang, Junhong Zhao, Yun Zhang, and Stefanie Zollmann.
\newblock A survey on 360° images and videos in mixed reality: Algorithms and applications.
\newblock \emph{Journal of Computer Science and Technology}, 38\penalty0 (3):\penalty0 473--491, 2023{\natexlab{a}}.

\bibitem[Zhang et~al.(2023{\natexlab{b}})Zhang, Yi, Jia, Wang, and Odamaki]{zhang2023panopoint}
Hengzhi Zhang, Hong Yi, Haijing Jia, Wei Wang, and Makoto Odamaki.
\newblock Panopoint: Self-supervised feature points detection and description for 360deg panorama.
\newblock In \emph{CVPR Workshops}, 2023{\natexlab{b}}.

\bibitem[Zhao et~al.(2015)Zhao, Feng, Wan, and Zhang]{zhao2015sphorb}
Qiang Zhao, Wei Feng, Liang Wan, and Jiawan Zhang.
\newblock Sphorb: A fast and robust binary feature on the sphere.
\newblock \emph{International journal of computer vision}, 113:\penalty0 143--159, 2015.

\bibitem[Zhou et~al.(2021)Zhou, Sattler, and Leal-Taixe]{zhou2021patch2pix}
Qunjie Zhou, Torsten Sattler, and Laura Leal-Taixe.
\newblock Patch2pix: Epipolar-guided pixel-level correspondences.
\newblock In \emph{CVPR}, 2021.

\bibitem[Zioulis et~al.(2018)Zioulis, Karakottas, Zarpalas, and Daras]{zioulis2018omnidepth}
Nikolaos Zioulis, Antonis Karakottas, Dimitrios Zarpalas, and Petros Daras.
\newblock Omnidepth: Dense depth estimation for indoors spherical panoramas.
\newblock In \emph{ECCV}, 2018.

\end{thebibliography}
}


\end{document}